\RequirePackage{fix-cm}
\documentclass{svjour3}                     
\smartqed  
\usepackage[square,numbers]{natbib}
\usepackage{amsmath}
\usepackage{graphicx}
\usepackage[utf8]{inputenc} 
\usepackage[T1]{fontenc}    
\usepackage{hyperref}       
\usepackage{url}            
\usepackage{booktabs}       
\usepackage{amsfonts}       
\usepackage{nicefrac}       
\usepackage{microtype}      
\usepackage{algorithm}
\usepackage[noend]{algpseudocode}
\usepackage{multirow}
\usepackage{caption}
\usepackage{subcaption}
\usepackage{float}
\usepackage{varwidth}
\begin{document}

\title{An adaptive synchronization approach for weights of deep reinforcement learning
}


\author{S.~Amirreza Badran          \and
        Mansoor Rezghi 
}


\institute{A. Badran \at
              Department of Computer Science \\
              Tarbiat Modares University\\
              \email{a.badran@modares.ac.ir}           
           \and
          Corresponding Author: M. Rezghi \at
              Department of Computer Science \\
              Tarbiat Modares University\\
              \email{rezghi@modares.ac.ir}  
}

\date{Received: date / Accepted: date}

\maketitle

\begin{abstract}
Deep Q-Networks (DQN) is one of the most well-known methods of deep reinforcement learning, which uses deep learning to approximate the action-value function.  Solving numerous Deep reinforcement learning challenges such as moving targets problem and the correlation between samples are the main advantages of this model. Although there have been various extensions of DQN in recent years, they all use a similar method to DQN to overcome the problem of moving targets. Despite the advantages mentioned, synchronizing the network weight in a fixed step size, independent of the agent's behavior, may in some cases cause the loss of some properly learned networks.  These lost networks may lead to states with more rewards, hence better samples stored in the replay memory for future training. 
In this paper, we address this problem from the DQN family and provide an adaptive approach for the synchronization of the neural weights used in DQN. In this method, the synchronization of weights is done based on the recent behavior of the agent, which is measured by a criterion at the end of the intervals. 
To test this method, we adjusted the DQN and rainbow methods with the proposed adaptive synchronization method. We compared these adjusted methods with their standard form on well-known games, which results confirm the quality of our synchronization methods.
\end{abstract}

\section{Introduction}
Reinforcement learning is a branch of machine learning that deals with how an agent can act in an environment to receive the maximum rewards from the environment.  
One of the well known and first successful neural networks used in reinforcement learning is \citet{tesauro1994td}~'s TD-gammon \citep{tesauro1994td}. It used neural networks to train the agent who won the world backgammon.
After that, one of the most significant breakthroughs in RL was the Deep Q network\cite{mnih2015human}, which used deep learning in the modeling of the action function.
They used convolutional neural networks in combination with experience replay so that the ‌ agent could learn from raw pixels of Atari games without any handcrafted features nor any changes for a specific game. Their experiments have shown that the DQN‌ agent could play Atari 2600 as good as a human.

In addition to using the deep neural networks, addressing the moving targets problem by enabling the network to be synced after a while rather than immediate updates, was another success factor of this method.  Their approach for synchronization increased the stability and possibility of convergence to the optimal policy compared to immediately updating the target network.

Following the success of DQN, many extensions of this method presented in recent years. For example, double DQN \cite{van2016deep}  addresses the overestimation bias of Q-learning by separating of selection and evaluation of actions \cite{hasselt2010double}.
Prioritized experience replay proposed in \cite{schaul2015prioritized} improved experience replay of DQN. This improvement is done based on selecting important experiences more frequently. 

The dueling network architecture \cite{Wang2016Dueling} is another extension that tries to generalize
across actions by separately representing  state values
and action advantages to different neural networks.
In distributional Q-learning \cite{bellemare2017distributional}, the agent learns to approximate the distribution of returns instead of expected returns.

Recently in \cite{hessel2018rainbow}, all of these extensions were integrated into one agent called Rainbow. In the reported experiments, this integrated method achieves the state of the art performance.

In all methods similar to the DQN, syncing the deep networks is done at fixed predefined intervals. Although this approach could solve the moving target problem, losing some semi-optimal trained target networks in some cases is its main drawback. Syncing the target network weights without considering the agent's behavior may prevent the agent from following a good policy which could lead to more potential rewards.  In this paper, we propose an adaptive way of syncing with considering the behavior of the agent.

\section{Background}
\label{sec:Background}

Reinforcement learning  problems can be formalized by the Markov Decision Process (MDP). MDP is a tuple $\langle \mathcal{S},\mathcal{A},T,r,\gamma \rangle$ where $\mathcal{S}$ is a finite set of states, $\mathcal{A}$ is a finite set of actions, $T$ is the transition matrix, $r$ is the reward function and $\gamma$ is a discount factor.
Usually, in reinforcement learning, an agent starts at state $S_1$ and then by selecting an action $A_t$, the environment sends a reward signal $R_1$ and the agent moves to the next state $S_{2}$ based on $T(S_1,A_t)$ and so on, until it reaches to a terminal state or the maximum number of steps allowed. 
\subsection{Basics of reinforcement learning}
Policy, value functions, and the model are relevant concepts in RL.

While the policy is the agent's behavior function, the value functions determine the quality of each state or each action. The model is the agent's representation of the environment. 
Temporal-Difference learning \cite{sutton1988learning} is One of the most important and widely used techniques in reinforcement learning.  In this method, the agent learns to predict the expected value of the value function at the end of each step. Expressed formally: 
\begin{equation}
V(S_t) \leftarrow V(S_t) + \alpha [R_{t+1}+\gamma V(S_{t+1})-V(S_t)]. \label{eqn:TDlearning}
\end{equation}
	On-policy and off-policy learning methods are two categories that used in RL. In on-policy learning the value functions us learned from the actions, the agent achieves using the current policy. In off-policy methods, the policy used to generate behavior may be unrelated to the policy that is evaluated and improved.
	
DQN (based on Q-learnin)g \cite{watkins1992q}  is an off-policy control algorithm of TD(0). In this algorithm, first action-values are initialized arbitrary (e.\ g. 0) and then in each step old value $Q(s_t,a_t)$ is updated by a fraction (learning rate)  of
$r+\gamma \max_{a'}Q^*(s',a')$ which called is the td-target $\delta$ \eqref{eqn:qlearning}. Q-learning is formulated as:
\begin{equation}
Q(S_t,A_t) \leftarrow Q(S_t,A_t) + \alpha \cdot \delta. \label{eqn:qlearning}
\end{equation}
where $\delta$ is td-target 
\begin{equation}
\delta = [R_{t+1}+\gamma \max_a Q(S_{t+1},a)-Q(S_t,A_t)]	
\end{equation}
and $\alpha$ is the learning rate.

Q-learning is considered a tabular solution method \cite{sutton2018reinforcement}. There are other types of RL methods that use approximation instead of a tabular solution. In Q-learning, we can approximate action-value function $Q$ instead of calculating the exact value of it. This is the main idea of deep reinforcement learning discussed in section \ref{sec:DQN}.
\subsection{Deep reinforcement learning}
\label{sec:DQN}
The authors in\cite{mnih2015human} developed a deep reinforcement learning method by using a deep convolutional neural network to approximate the action-value function. Using CNN architecture enabled them to learn from raw images (in this instance, images of Atari 2600 games) without any handcrafted features.  Furthermore, the same settings and network architecture used from variant games and DQN showed the state of the art performance in the majority of the games.  The development of DQN was challenging, one of which was moving targets problem. In each iteration, the values of Q may change, thus altering the value of the targets, which may cause oscillations or divergence.  In  \citet{mnih2015human}, to overcome this challenge the weights of the action value network for C updates are cloned and used as the Q-learning updates.
 The duplicated Q-network is showed by $\hat{Q}$.
Another challenge of using DQN was the correlation between successive samples, which led to a large variation between updates, thus instability of the method. All of mentioned extensions (including Rainbow) are using the same approach as DQN to overcome the moving targets problem. \citet{mnih2015human} Modified the standard Q-learning to use experience replay \cite{lin1992self}. Experiences $e_t=(s_t,a_t,r_t,s_{t+1})$ at each time-step $t$ are stored in a memory data set $D_t=\{e_1,\cdots,e_t\}$ and during learning Q-learning updates will be applied by samples of these experiences which are selected randomly with a uniform distribution. 

The raw images of Atari games are preprocessed by function $\phi$ (described in \cite{mnih2015human}) and turned into an $84\times84\times4$ image (or tensor). Q-network consists of three convolutional layers and two fully connected layers with rectifier nonlinearity (that is, $\max(0,x)$) as activation function and the output layer has a  single output for each valid action. The Q-network can be trained by optimizing the parameters $\theta_i$ at iteration $i$ to reduce the mean-squared error in the Bellman equation, where the optimal target values $r+\gamma \max_{a'}Q^*(s',a') $ are substituted with approximate target values $y= r+\gamma \max_{a'}Q^*(s',a';\theta_i^{-})$, using parameters $\theta_i$ from some previous iteration. This leads to a sequence of loss functions $L_i(\theta_i)$ that changes at each iteration i,
\begin{equation}
L_i(\theta_i) = \mathbb{E}_{s,a,r,s'\sim D_i }\bigg[(\dot{\delta} - Q(s,a;\theta_i))^2\bigg]
\label{eqn:lossfunction}
\end{equation}
where
\begin{equation}
\dot{\delta} = r+\gamma \max_{a'}Q^*(s',a';\theta_i^{-})
\end{equation}
Which can be optimized by gradient descent methods, specifically stochastic  gradient descent  method.

\begin{algorithm}
	\caption{deep Q-learning with experience replay.  
	} \label{alg:DQN}
	\begin{algorithmic}[1]
		\State Initialize replay memory $ D $ to capacity $ N $
		
		\State Initialize action-value function $ Q $ with random weights $ \theta $
		
		\State Initialize target action-value function $ \hat{Q} $ with weights $ \theta^{-}=\theta $
		
		\For{$episode = 1, \cdots, M$}
		
		\State \begin{varwidth}[t]{0.9\linewidth}
			Initialize sequence $s_1 = \{x_1\}$ and preprocessed sequence $\phi_1=\phi(s_1)$
		\end{varwidth}
		\For{$t=1, \cdots, T$}
		\State With probability $\epsilon$ select a random action  $a_t$
		
		\State otherwise select $a_t = argmax_a Q(\phi(s_t),a;\theta)$
		
		\State \begin{varwidth}[t]{0.8\linewidth}
			Execute action $a_t$  in emulator and observe reward $r_t$ and image 	$x_{t+1}$
		\end{varwidth}
		\State \begin{varwidth}[t]{0.8\linewidth}
			set $s_{t+1}=(s_t,a_t,x_{t+1})$  and preprocess \mbox{$\phi_{t+1}=\phi(s_{t+1})$}
		\end{varwidth}
		\State Store transition 	$(\phi_t,a_t,r_t,\phi_{t+1})$ in $D$ 
		
		\State \begin{varwidth}[t]{0.8\linewidth}
			Sample random minibatch of transitions 		$(\phi_j,a_j,r_j,\phi_{j+1})$ from $D$
		\end{varwidth}
		\State \begin{varwidth}[t]{0.8\linewidth}
			\[
			\textnormal{Set }y_j=\Bigg\{\begin{array}{lc}
			r_j, &  s_{j+1} \textnormal{is terminal }\\ 
			\dot{\delta}, & \textnormal{o.w.}
			\end{array}
			\]
			where $\dot{\delta} = r_j+\gamma \max_{a'}\hat{Q}(\phi_{j+1},a';\theta^{-})$
		\end{varwidth}
		\State \begin{varwidth}[t]{0.8\linewidth}
			Perform a gradient descent step on $(y_j - Q(\phi_j,a_j;\theta))^2$ with respect to  $\theta$
		\end{varwidth}
		\State Every $C$ steps reset		$\hat{Q}=Q$
		
		\EndFor
		\EndFor
		
	\end{algorithmic}
\end{algorithm}

\section{Proposed adaptive synchronization method}
\label{sec:ProposedMethod}
As mentioned, in DQN and it's extensions, syncing the deep networks is done at intervals with the fixed step size $C$ to solve the moving target problem. 
Syncing the target network weights in a fixed step size $C$ without considering the agent's behavior may prevent the agent from following a good policy and leading to more potential rewards.  In this section,  we define a criterion to find out when to sync the network instead of the fixed steps size.
  This criterion should set up based on the obtained rewards by the agent.  Since these rewards are obtained by the actions of the agent(chosen by   $a_t = argmax_a \hat{Q}(s_t, a;\theta)$) then they are dependent to the Q-network that is learning continuously by target network (see \eqref{eqn:lossfunction}).  Therefore,  we can use a stream of recent rewards to measure the behavior of the agent as an evaluation of the network quality. Then, based on this evaluation criterion, we can decide when to synchronize the target network with the Q-network.

Let  $t$ be a step that we should decide about syncing. To construct our criterion for syncing, we push the  recent rewards (obtained from the environment) in a  queue 
$ \{r_{2n-t+1}, \ldots, r_t\}$
with the length $2n$  and compare the  quality of rewards of the first part, i.e,
$que_{old}  =\{ r_{t-2n+1}, \cdots, r_{t-n}\}$
and the second part 
$que_{new}  =\{ r_{t-n+1}, \cdots, r_t\}$
of this queue to construct our criterion. Here $n$ could be selected as $C$. If the second part works better than the first one this means that the agent and so the network works good and we should preserve this network.  

Average of rewards can tell us about the behavior of agent but since we are more interested in recent behavior,  applying a weight vector $w$ which increases the effect of recent rewards  can lead to a better representation of recent  behavior of the agent.  So this weight vector should satisfy the following conditions
\begin{align}
w_{1}\leq w_2 \leq \ldots \leq w_n, \quad \sum_{i=1}^{n}w_i=1
\end{align}
Although different type of weights could be used, here we use a simple form
\[
w_{i}=\frac{i}{n(n+1)}.
\] 
So the weighted average of rewards in both $que_{new}$ and $que_{old}$ will be
\begin{align}
\overline{que}_{new} &= {\frac{1}{n}}{\sum\limits_{i = 1}^{n} w_i (que_{new})_i} \label{eqn:AVGQue1}\\
\overline{que}_{old} &= {\frac{1}{n}}{\sum\limits_{i = 1}^{n} w_i (que_{old})_i} \label{eqn:AVGQue2}
\end{align}
Now we need to decide whether to sync $\hat{Q}=Q$ or not, for this matter, we used the simple idea of comparing the weighted average of $que_{new}$ and $que_{old}$. 
We, at each $C$ steps, calculate $\overline{que}_{new}$ and $\overline{que}_{old}$ (\eqref{eqn:AVGQue1}  , \eqref{eqn:AVGQue2}) and if $\overline{que}_{new} < \overline{que}_{old}$ which means the recent behavior of the agent is not as good as last time, we sync the weights of Q-network. Because of adaptability of our synchronization method, we call our modified DQN agent the Adaptive Synchronization DQN or for short AS\_DQN and our modified Rainbow agent, AS\_Rainbow.

\begin{algorithm}
	\caption{Adaptive Synchronization DQN}
	\label{alg:ASDQN}
	\begin{algorithmic}[1]

		\State Initialize replay memory $ D $ to capacity $ N $
		
		\State Initialize action-value function $ Q $ with random weights $ \theta $
		
		\State Initialize target action-value function $ \hat{Q} $ with weights $ \theta^{-}=\theta $
		
		\For{$episode = 1, \cdots, M$}
		
		\State \begin{varwidth}[t]{0.9\linewidth}
			Initialize sequence $s_1 = \{x_1\}$ and preprocessed sequence $\phi_1=\phi(s_1)$
		\end{varwidth}
		\State \begin{varwidth}[t]{0.9\linewidth}
			Initialize $que$  a $2n$ queue containing the rewards with weight vector $w$.
			
		\end{varwidth}
		\For{$t=1, \cdots, T$}
		\State With probability $\epsilon$ select a random action  $a_t$
		
		\State otherwise select $a_t = argmax_a Q(\phi(s_t),a;\theta)$
		
		\State \begin{varwidth}[t]{0.85\linewidth}
			Execute action $a_t$  in emulator and observe reward $r_t$ and image 	$x_{t+1}$
		\end{varwidth}
		\State \begin{varwidth}[t]{0.85\linewidth}
			set $s_{t+1}=(s_t,a_t,x_{t+1})$  and preprocess 		$\phi_{t+1}=\phi(s_{t+1})$
		\end{varwidth}
		\State \begin{varwidth}[t]{0.85\linewidth}
			Store transition 	$(\phi_t,a_t,r_t,\phi_{t+1})$ in $D$ and push $r_t$ into $que$ 
		\end{varwidth}
		\State \begin{varwidth}[t]{0.85\linewidth}
			Sample random minibatch of transitions 		$(\phi_j,a_j,r_j,\phi_{j+1})$ from $D$
		\end{varwidth}
		\State \begin{varwidth}[t]{0.85\linewidth}
			\[
			\textnormal{Set }y_j=\Bigg\{\begin{array}{lc}
			r_j, &  s_{j+1} \textnormal{is terminal }\\ 
			\dot{\delta}, & \textnormal{o.w.}
			\end{array}
			\]
			where $\dot{\delta} = r_j+\gamma \max_{a'}\hat{Q}(\phi_{j+1},a';\theta^{-})$
		\end{varwidth}
		\State \begin{varwidth}[t]{0.85\linewidth}
			Perform a gradient descent step on $(y_j - Q(\phi_j,a_j;\theta))^2$ with respect to  $\theta$
		\end{varwidth}
		\If{$mod(step,C)==0$}
		\State
		$\overline{que}_{new} = {\frac{1}{n}}{\sum\limits_{i = 1}^{n} w_i (que_{new} )_i}$
		\State
		$\overline{que}_{old} = {\frac{1}{n}}{\sum\limits_{i = 1}^{n} w_i (que_{old} )_i}$
		
		\If{$\overline{que}_{new} < \overline{que}_{old}$}
		\State
		Reset		$\hat{Q}=Q$
		\EndIf
		\EndIf
		
		\EndFor
		\EndFor
		
	\end{algorithmic}
\end{algorithm}
\section{Experimental results}
\label{ExperimentalResults}
In this section, we describe our settings and configurations used to test our agents. We used a Linux-based PC to train our agents, and each session lasts between 3 and 4 days. For DQN implementation, we used the Dopamine \cite{r51} library and modified their DQN and Rainbow implementations with our proposed method. For the testing environment, we used Atari 2600 games provided by OpenAi's Gym \cite{r53}.   

\subsection{Evaluation methodology}
We did the experiments on   AirRaid, Alien,  Amidar, Assault, Asterix, Asteroids, Breakout, SpaceInvaders, and Seaquest games. We tried to choose games that are different in visuals and gameplay.   In our all experiments all parameters (even the target network period  $C$ ) were same as DQN \cite{mnih2015human} and Rainbow \cite{hessel2018rainbow} accordingly and for our hyper-parameters we tested our agent with $k = 3$ and $\delta = 0.01$ .Also, similar to \cite{mnih2015human} and \cite{hessel2018rainbow} we trained our agents with 200 iterations, where each iteration consists of two training and evaluation phases.  In the training phase, the agent is trained on 1M frames, and in the evaluation phase, the agent is evaluated only on 512k frames.  In both phases, we used the mean of episode return (scores) to evaluate our agents and compare them to DQN and Rainbow. For DQN and Rainbow results, we used data prepared by Dopamine \ cite {r51}. Also, the reported results for each game are an average of 5 training sessions. Although due to limited computing tools and time, we only trained our agent for 3 sessions in 4 games, in all 9 games, our agents were even better than the best DQN representatives. 

\subsection{Analysis}
\begin{figure}
	\centering
	\includegraphics[width=\columnwidth]{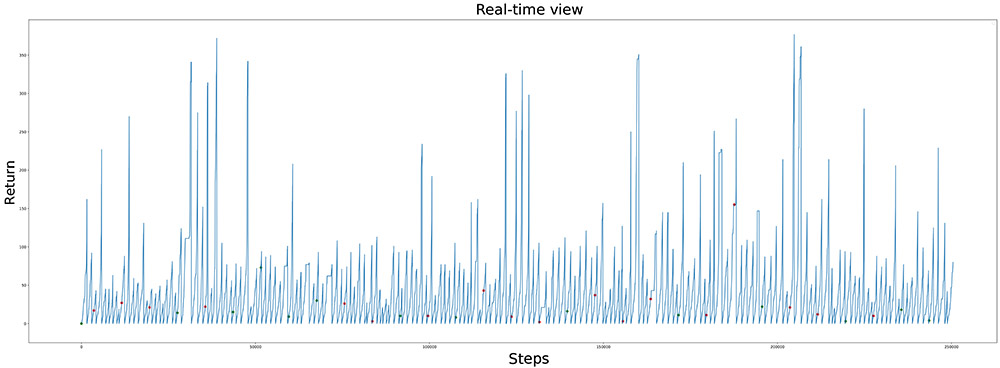}
	\caption{Real-time return of AS\_DQN agent after 50 iterations of training. Red marks mean the agent did not sync its target network and green marks mean it did.}
	
	\label{fig:ASDQNrealtime}
	
\end{figure}

\begin{figure}
	\centering
	\begin{subfigure}[t]{0.75\columnwidth}
		\includegraphics[width=\columnwidth]{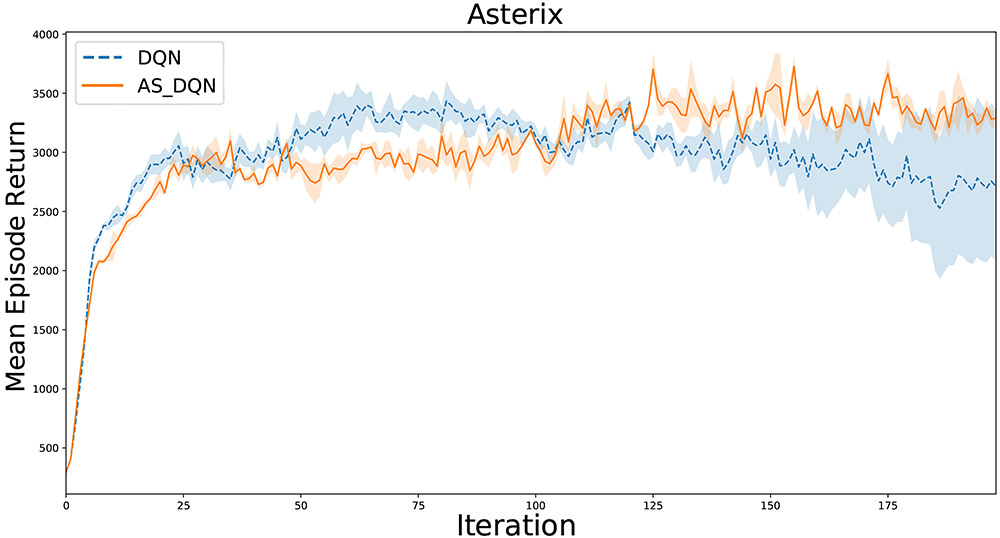}
		\caption{Asterix}
		\label{fig:Asterix_DQN}
	\end{subfigure}%
	\quad
	\begin{subfigure}[t]{0.75\columnwidth}
		\includegraphics[width=\columnwidth]{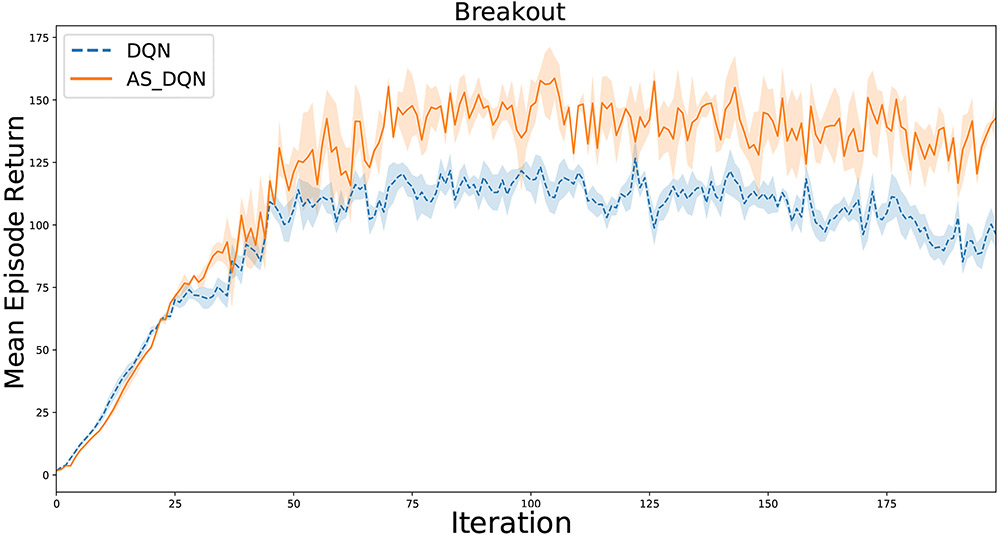}
		\caption{Breakout}
		\label{fig:Breakout_DQN}
	\end{subfigure}%
	\\
	\begin{subfigure}[b]{0.75\columnwidth}
		\includegraphics[width=\columnwidth]{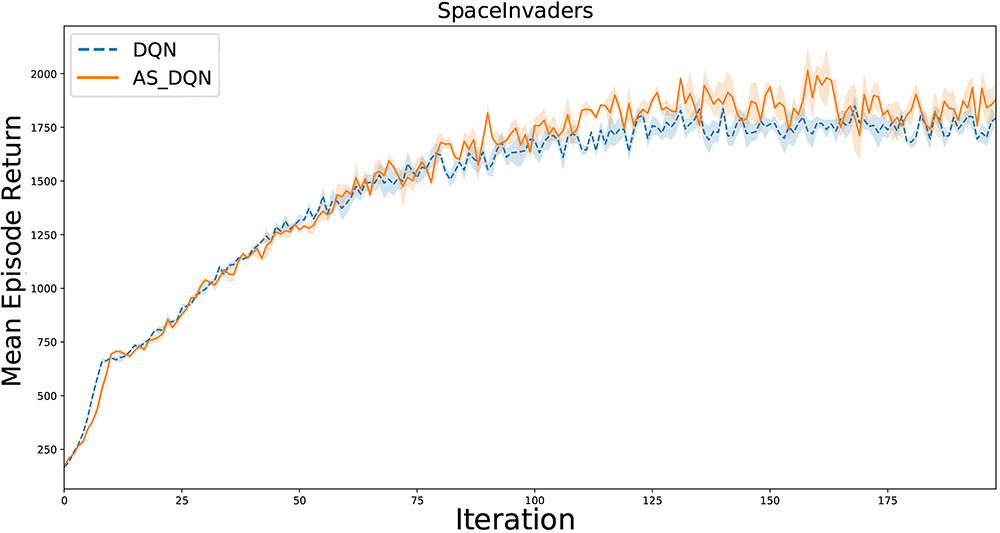}
		\caption{SpaceInvaders}
		\label{fig:SpaceInvaders_DQN}
	\end{subfigure}%
	\quad
	\begin{subfigure}[b]{0.75\columnwidth}
		\includegraphics[width=\columnwidth]{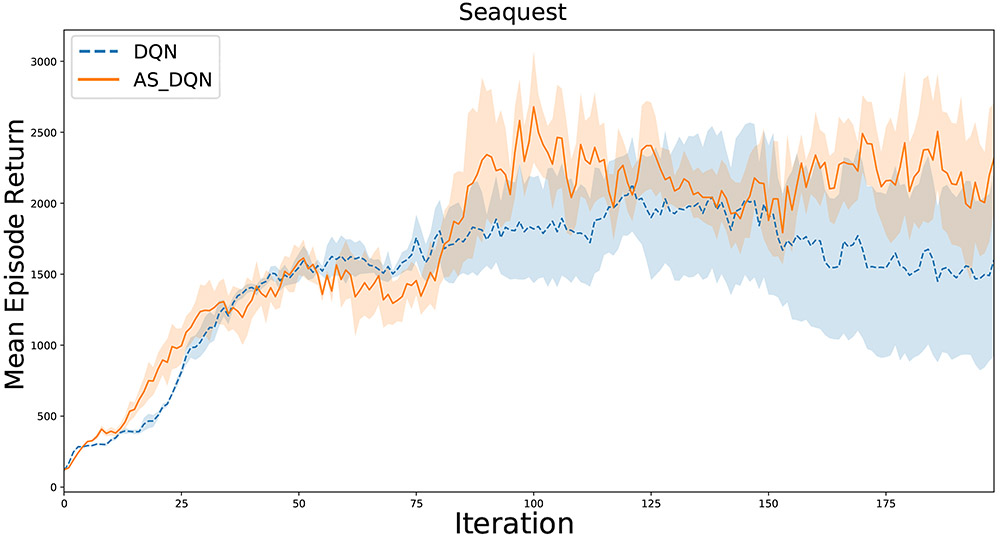}
		\caption{Seaquest}
		\label{fig:Seaquest_DQN}
	\end{subfigure}%
	\caption[ASDQN]{ \textbf{DQN vs AS\_DQN}.  Average return of episodes for each iteration are shown for 4 different games. DQN results are averaged over 5 runs and AS\_DQN runs are averaged over 3 runs in this figure.  }
	
	\label{fig:ASDQNresults}
	
\end{figure}

\begin{figure}
	\centering
	\begin{subfigure}[t]{0.6\columnwidth}
		\includegraphics[width=\columnwidth]{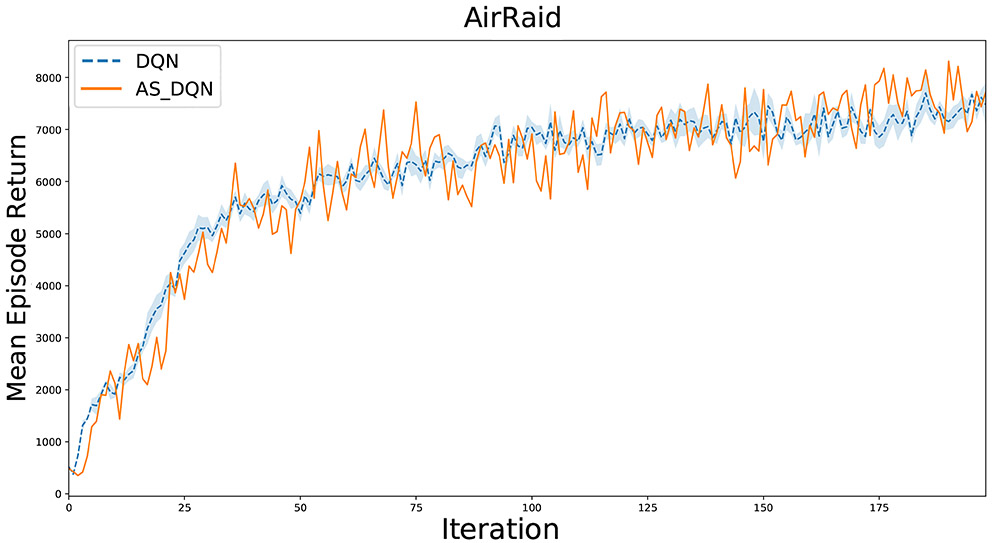}
		\caption{AirRaid}
		\label{fig:ASDQN_AirRaid}
	\end{subfigure}%
	\quad
	\begin{subfigure}[t]{0.6\columnwidth}
		\includegraphics[width=\columnwidth]{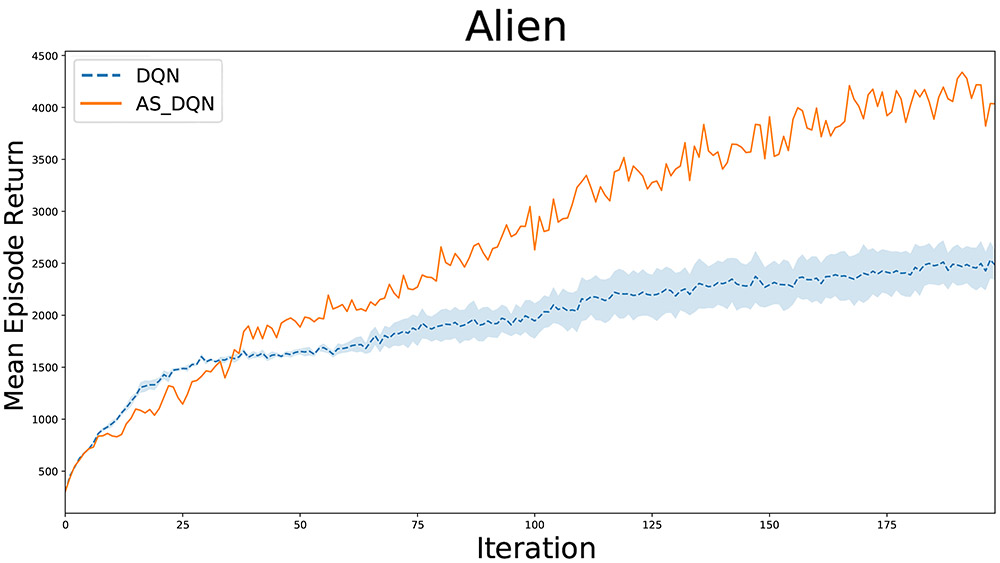}
		\caption{Alien}
		\label{fig:ASDQN_Alien}
	\end{subfigure}%
	\\
	\begin{subfigure}[b]{0.6\columnwidth}
		\includegraphics[width=\columnwidth]{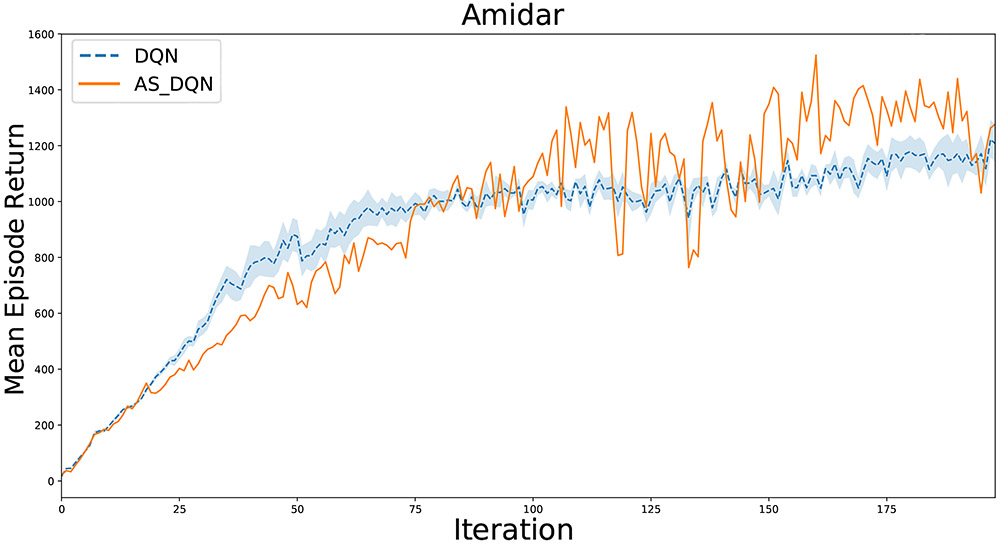}
		\caption{Amidar}
		\label{fig:ASDQN_Amidar}
	\end{subfigure}%
	\quad
	\begin{subfigure}[b]{0.6\columnwidth}
		\includegraphics[width=\columnwidth]{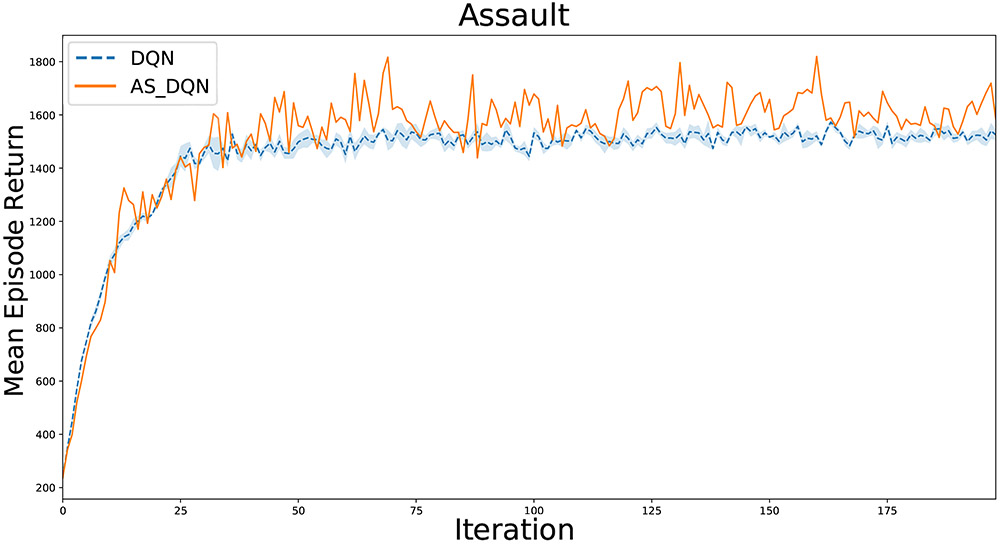}
		\caption{Assault}
		\label{fig:ASDQN_Assault}
	\end{subfigure}%
	\quad
	\begin{subfigure}[b]{0.6\columnwidth}
		\includegraphics[width=\columnwidth]{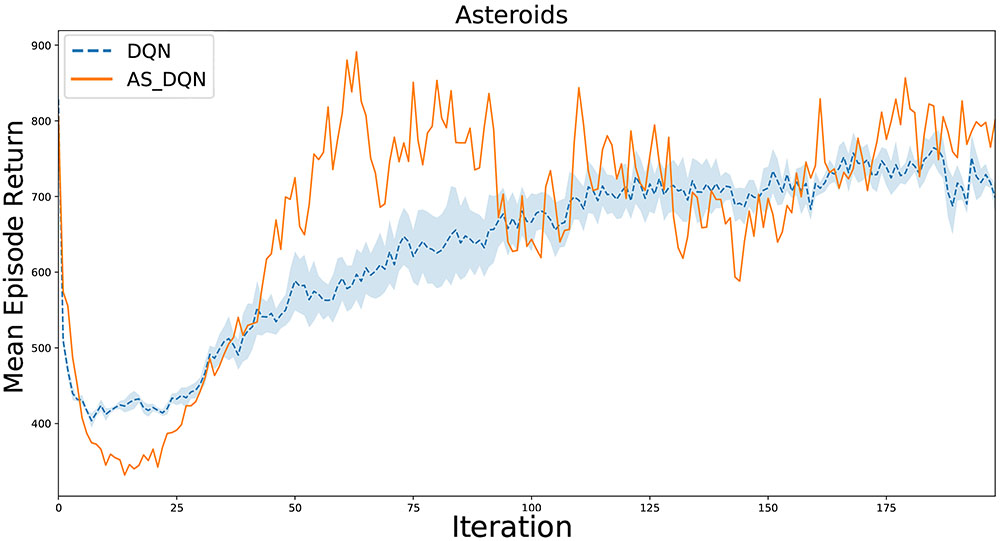}
		\caption{Asteroids}
		\label{fig:ASDQN_Asteroids}
	\end{subfigure}%
	\caption[ASDQN2]{ \textbf{DQN vs AS\_DQN}. Average return of episodes for each iteration are shown for 5 different games. DQN results are averaged over 5 runs and AS\_DQN runs are averaged over only one in this figure.}
	
	\label{fig:ASDQNresults2}
	
\end{figure}

First, we trained our agent on the Breakout game for 50 iterations. Then we plot the real-time return for 1 million frames (or 250k steps) and marked when our agent synced its target network (figure \ref{fig:ASDQNrealtime}). The agent behavior was as expected. It maintained the target networks which it achieved higher returns and synced to new ones when the performance did not meet the criteria.\\ 
As mentioned before, we tested our agents on 9 Atari 2600 games. We used DQN baselines provided by Dopamine library \cite{r51}. AS\_DQN outperformed DQN in almost all 9 games. To be more specific, the average rewards of 3 AS\_DQN agents outperformed the average of 5 DQN agent on 4 games (figure \ref{fig:ASDQNresults}). Also, AS\_DQN agents had less variance over different runs (e.g. the variance of DQN returns over 5 runs of Seaquest is 916200 compared to AS\_DQN's 590507). Furthermore, our single AS\_DQN agents outperformed the average of 5 DQN agents on 5 other games (figure \ref{fig:ASDQNresults2}).

\begin{figure}
	\centering
	\begin{subfigure}[t]{0.75\columnwidth}
		\includegraphics[width=\columnwidth]{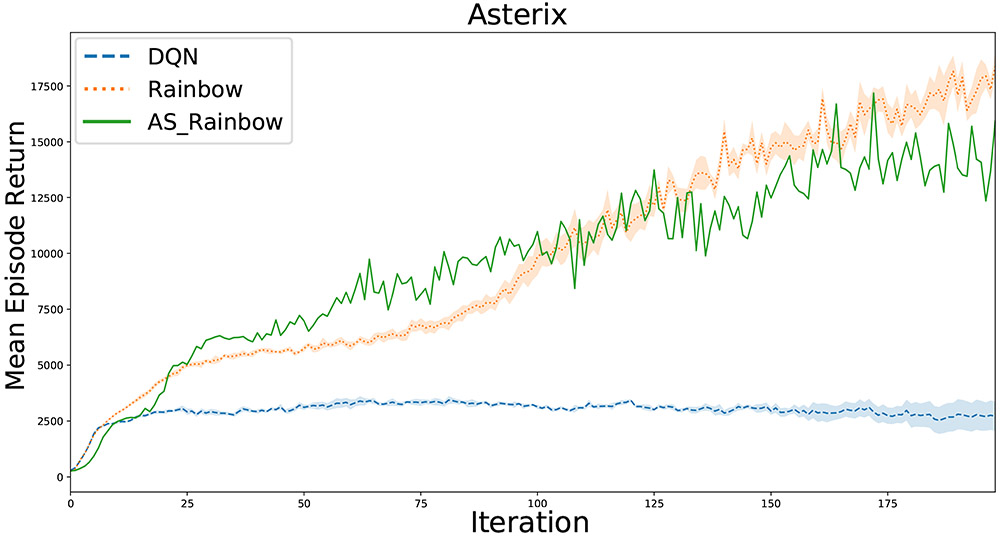}
		\caption{Asterix}
		\label{fig:Asterix_Rainbow}
	\end{subfigure}%
	\quad
	\begin{subfigure}[t]{0.75\columnwidth}
		\includegraphics[width=\columnwidth]{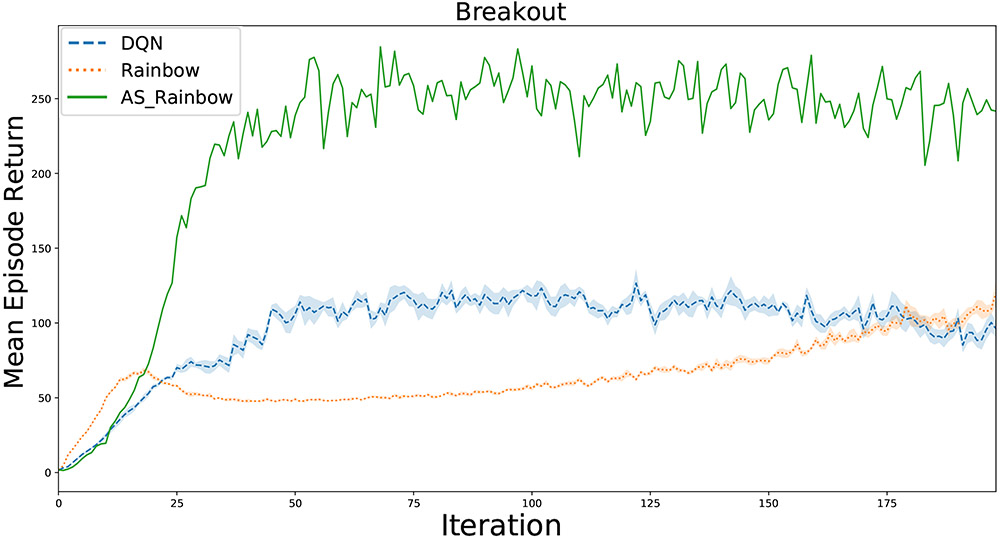}
		\caption{Breakout}
		\label{fig:Breakout_Rainbow}
	\end{subfigure}%
	\\
	\begin{subfigure}[b]{0.75\columnwidth}
		\includegraphics[width=\columnwidth]{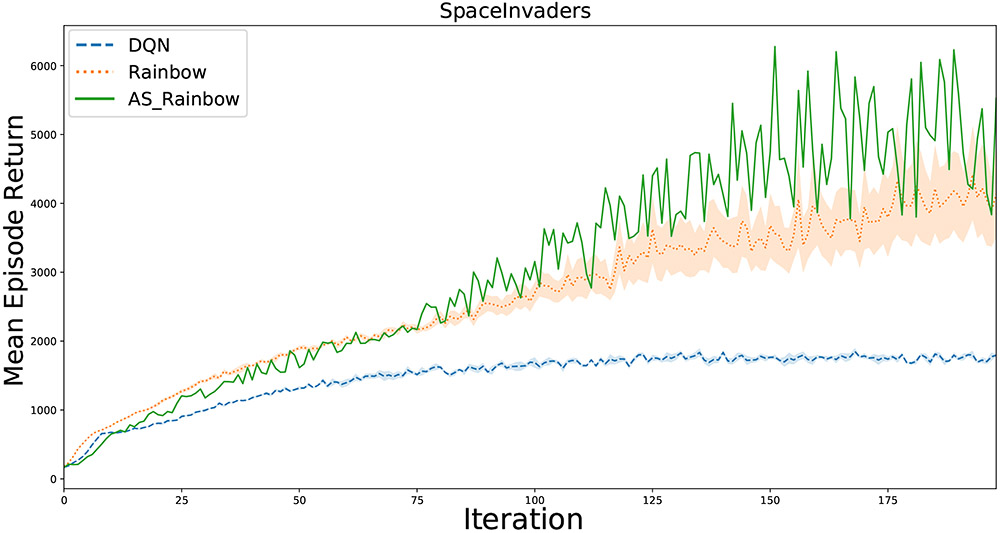}
		\caption{SpaceInvaders}
		\label{fig:SpaceInvaders_Rainbow}
	\end{subfigure}%
	\quad
	\begin{subfigure}[b]{0.75\columnwidth}
		\includegraphics[width=\columnwidth]{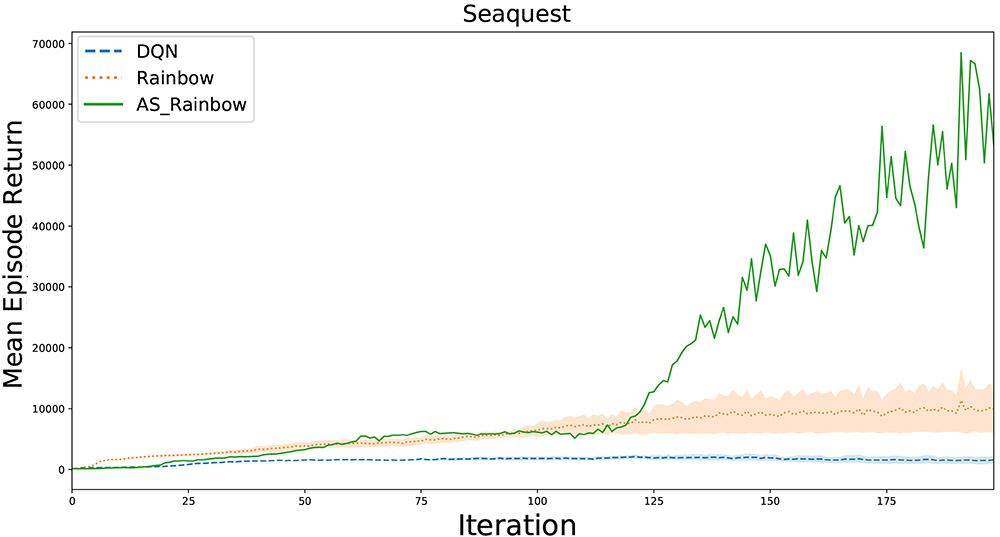}
		\caption{Seaquest}
		\label{fig:Seaquest_Rainbow}
	\end{subfigure}%
	\caption[AS\_Rainbow]{ \textbf{Rainbow vs AS\_Rainbow}.  Average return of episodes for each iteration are shown for 4 different games. DQN results  are also shown for better insight of Rainbow and AS\_Rainbow performances.}
	
	\label{fig:ASRainbowresults}
	
\end{figure}

AS\_Rainbow shows an exceptionally good performance in Breakout and Seaquest (Figures \ref{fig:Breakout_Rainbow} and \ref{fig:Seaquest_Rainbow}) but Rainbow outperformed AS\_Rainbow in Asterix (Figure \ref{fig:Asterix_Rainbow}) and AS\_Rainbow performs better than Rainbow in SpaceInvaders (Figure \ref{fig:SpaceInvaders_Rainbow}).

\begin{figure}
	\centering
	\begin{subfigure}[t]{0.75\columnwidth}
		\includegraphics[width=\columnwidth]{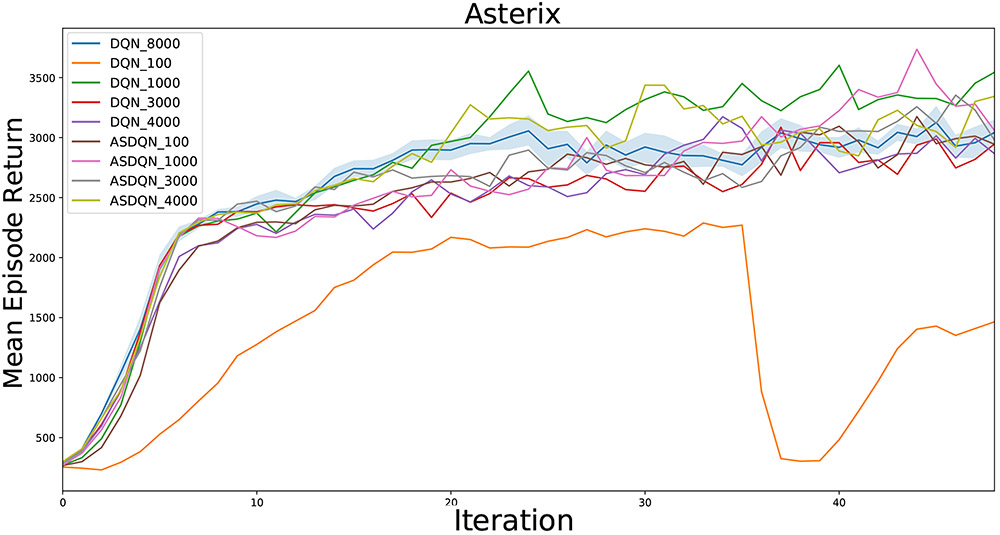}
		\caption{Asterix}
		\label{fig:Asterix_ASDQNsmallsteps}
	\end{subfigure}%
	\quad
	\begin{subfigure}[t]{0.75\columnwidth}
		\includegraphics[width=\columnwidth]{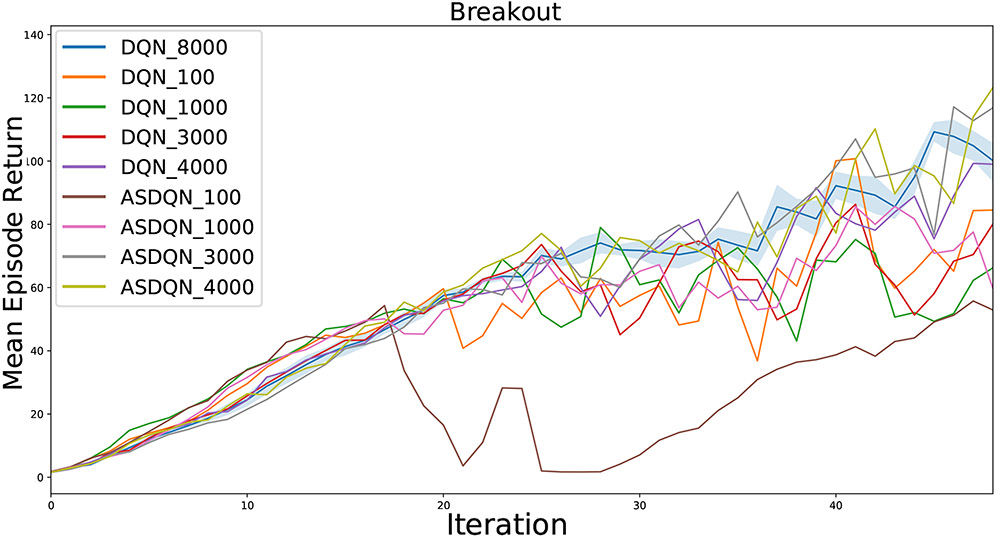}
		\caption{Breakout}
		\label{fig:Breakout_ASDQNsmallsteps}
	\end{subfigure}%
	\caption[AS\_DQN small steps]{ \textbf{DQN vs AS\_DQN}.  Average return of episodes for each iteration are shown for 2 different games with $C=100,1000,3000,4000$ and $C=8000$ as benchmark.}
	
	\label{fig:ASDQNsmallsteps}
	
\end{figure}

The main problem in methods that use the target network is finding the optimal length of fixed intervals. If we choose a very small step size $C$ for fixed intervals, it leads to instability during training the Q-network, which is not desirable(see DQN agent with $C=100$ in figure \ref{fig:ASDQNsmallsteps}); And if we choose a very long  step size $C$, it may lead to overfitting  the Q-network to every target network, which again, is not desirable. On the other hand, scale of these step sizes is different for every problem and environment, as you may already know, Atari's Pong game can be trained by a small step size (e.g. $C=100$) which, if used for some other game (e.g. SpaceInvaders) leads to divergence from optimal policy. Therefor, finding appropriate step size  is an important problem for DQN and methods based on it and considering only a predefined step size  is not optimal. Despite this, due to our knowledge all state of the art methods like Rainbow based on DQN are using a fixed step size. To overcome this problem the main question is that how we define a criterion for quality of a target network? Is the instant rewards that agent obtains from environment a good  criterion? These rewards are obtained by  actions of the agent, which it chooses based on the current Q-network that is trained by target network. If we use a moving average over recent rewards we can eliminate any instant reward noises and have an acceptable (based on results)  criterion.

At last we desire to sync the target network as soon  as possible, but in fixed step sized method, networks sync at a predetermined rate. In our method if we choose smaller step sizes, the agent tend to sync the networks as soon as the criteria  allows it, and thats why our agent performs better at step sizes lower than $C=8000$ (figure \ref{fig:ASDQNsmallsteps}).

\section{Conclusion}

In this paper,  we investigated the DQN Reinforcement learning methods and considered its approach to solve moving targets problem. Here we discussed the drawback of syncing the network in fixed step size used by DQN and its extension, then shown that by evaluating the behavior of agent one could propose an adaptive syncing method.
Optimally we want to sync the target network as soon as possible, but by reducing the step size $C$ we risk the stability of the agent learning process. This type of adaptive approach discussed in this paper, might be the answer to this problem, no matter what is our environment (Atari or not), Our agents could have stable learning process towards the optimal policy without finding and fine tuning a fixed step size.    
For this matter, we proposed a simple method for this evaluation and experiments show the quality of this simple evaluation criteria. It is clear that in the future, we and others could derive more sophisticated
criterion for the quality of target networks.
In our test	we did not change any hyper-parameters which were used in \cite{mnih2015human} and \cite{hessel2018rainbow} to make a fair comparison. 
Due to time and hardware limitations our experiment results are not many but still enough to prove our method is effective.

	\bibliographystyle{spbasic}
	\bibliography{bibtexrefs}

\end{document}